\pdfoutput=1

\documentclass[11pt]{article}

\usepackage{acl}

\usepackage{times}
\usepackage{latexsym}
\usepackage[T1]{fontenc}
\usepackage[utf8]{inputenc}
\usepackage{microtype}

\usepackage{url}
\usepackage{booktabs}
\usepackage{graphicx}
\usepackage{amsmath}
\usepackage{tabularx}
\usepackage{ragged2e}
\usepackage{enumitem}
\usepackage{siunitx}
\usepackage{hyperref}

\usepackage{float}

\usepackage{tikz}
\usetikzlibrary{shapes.geometric, arrows}
\usetikzlibrary{shapes.geometric, arrows.meta, positioning, backgrounds, fit}

\newcolumntype{Y}{>{\centering\arraybackslash}X} 

\title{eSapiens's DEREK Module: Deep Extraction \& Reasoning Engine for Knowledge with LLMs}

\author{Isaac Shi \and Zeyuan Li \and Fan Liu \and Wenli Wang \\
        \bf{Lewei He \and Yang Yang \and Tianyu Shi} \\
  eSapiens Team \\
  \texttt{\{ishi, zeyuanli, fanliu, wenliwang, leweihe, yangyang, tianyushi\}@esapiens.ai} \\}

\begin{document}
\maketitle

\begin{abstract}
We present the \textbf{DEREK (Deep Extraction \& Reasoning Engine for Knowledge) Module}, a secure and scalable Retrieval-Augmented Generation pipeline designed specifically for enterprise document question answering. Designed and implemented by eSapiens, the system ingests heterogeneous content (PDF, Office, web), splits it into 1,000-token overlapping chunks, and indexes them in a hybrid HNSW+BM25 store. User queries are refined by GPT-4o, retrieved via combined vector+BM25 search, reranked with Cohere, and answered by an LLM using CO-STAR prompt engineering. A LangGraph verifier enforces citation overlap, regenerating answers until every claim is grounded.

On four LegalBench subsets, 1000-token chunks improve Recall@50 by $\approx1$~pp and hybrid+rerank boosts Precision@10 by $\approx7$~pp; the verifier raises TRACe Utilization above 0.50 and limits unsupported statements to $<3\%$. All components run in containers, enforce end-to-end TLS 1.3 and AES-256. These results demonstrate that the DEREK module delivers \emph{accurate, traceable}, and \emph{production-ready} document QA with minimal operational overhead.
The module is designed to meet enterprise demands for secure, auditable, and context-faithful retrieval, providing a reliable baseline for high-stakes domains such as legal and finance.
More details and demos are available at: \url{https://www.esapiens.ai/}.
\end{abstract}

\section{Introduction}
Enterprises manage vast collections of contracts, standard-operating procedures and engineering manuals that must be queried under strict compliance rules. Keyword search returns excessive noise, while large-language models (LLMs) hallucinate when the correct passage is missing. The \textbf{DEREK Module} closes this gap through a comprehensive pipeline. It begins by ingesting PDF, Office, and web content through one-click connectors. It then chunks and embeds these documents into 1,000-token overlapping windows, storing them in a hybrid index that blends HNSW vectors with BM25 term filters. Finally, it retrieves, reranks, and verifies the top-k snippets before passing them—along with a refined query—to an LLM that must cite every supporting span.

The core architecture is delivered as containerized services, ensuring that all content, embeddings, and logs are encrypted via HTTPS and AES-256. Every API call is checked against role-based policies, and audit events are persisted to satisfy SOC 2 traceability, allowing teams to scale the module with standard DevOps tooling.

\subsection*{Contributions}
Internal tests on four \emph{LegalBench} subsets were designed to confirm that the default settings shipped with the module are technically sound:

\begin{itemize}[leftmargin=*,itemsep=1pt]
  \item \textbf{Chunk size.} Moving from 500-token to 1000-token windows raises Recall@50 by roughly one percentage point without adding latency, validating 1000 tokens as the out-of-box default.
  \item \textbf{Hybrid search.} Adding a lightweight BM25 filter in front of the vector index removes obviously irrelevant passages and improves precision at low~$k$, giving the LLM cleaner context.
  \item \textbf{Verifier loop.} Enabling the \emph{LangGraph} verifier pushes TRACe Utilization above~0.5 and drives unsupported statements into the low single-digit range, demonstrating that the model truly grounds its answers in retrieved text.
\end{itemize}

These findings confirm that the shipped pipeline balances recall, precision and cost—providing a reliable baseline before any domain-specific tuning. The evaluation confirms that the module’s shipped architecture—1000-token chunks, hybrid search and mandatory verification—meets its design goal of \emph{secure, auditable and context-faithful retrieval}. Organisations can therefore pilot the module “as is”, adding only domain--specific documents or threshold adjustments, confident that the core pipeline behaves predictably under real workloads.

\section{Motivation}
\label{sec:motivation}
Although interest in Retrieval-Augmented Generation (RAG) is high, most organisations still struggle to obtain \emph{trustworthy} answers from their own document collections. Our design is motivated by several recurring technical challenges observed in enterprise deployments. They set the acceptance criteria for the DEREK Module.

\subsection{Fragmented Knowledge Bases}
Enterprise knowledge is typically balkanized. Unstructured content is scattered across PDFs, Office files, and SharePoint sites, while critical structured data is locked away in databases like PostgreSQL, Snowflake, or RDS. Each silo has its own authentication scheme and data model, preventing a single, unified query from accessing the full corpus. This creates an \textit{incomplete retrieval surface}: critical document passages and database records are invisible to traditional search and most open-source RAG pipelines.

\subsection{Recall–Precision Trade-off}
Pure vector search often misses semantically relevant yet lexically dissimilar passages; pure keyword search floods the LLM with noise. Hybrid retrieval can resolve the dilemma only if the system ranks vector similarity, term filters and metadata constraints in one coherent list. Most toolkits leave that orchestration to end users, producing uneven coverage and hard-to-debug failure modes.

\subsection{Hallucination and Missing Citations}
Even when the right snippet is retrieved, many pipelines simply paste it into a prompt and \emph{hope} the LLM behaves. Without a verifier loop that rejects unsupported claims, answers may “sound right” yet cannot be traced to source text. Such outputs fail internal audits in finance, healthcare and other high-stakes domains.

\subsection{Operational Overhead}
A production-grade RAG workflow spans ETL, chunking, embedding, hybrid indexing, cache invalidation, multi-agent orchestration, prompt version control and cost monitoring. Stitching these components together requires cross-stack expertise that is scarce outside big-tech. Maintenance costs can erase the very ROI that RAG was meant to deliver, turning many proofs of concept into shelf-ware.

\vspace{0.5em}
\noindent The remainder of this report explains how the DEREK Module is built to \emph{unify} siloed documents, \emph{balance} recall and precision, \emph{enforce} citation-based verification, and inherit platform-level security—while keeping operations lightweight.

\section{Related Work}
The emergence of Retrieval-Augmented Generation (RAG) has significantly improved the factuality and traceability of large language model (LLM) outputs in knowledge-intensive domains \citep{lewis2020retrieval, izacard2021distilling}. Early frameworks like RAG \citep{lewis2020retrieval} and FiD \citep{izacard2021distilling} focus on tightly coupling retrievers with generators to improve response accuracy, but they are primarily research prototypes with limited support for enterprise deployment concerns such as access control, versioning, or compliance.

In the product space, LangChain \citep{chase2022langchain} and LlamaIndex \citep{liu2022llamaindex} have popularized modular toolchains for building LLM-powered applications. These systems provide developer-centric orchestration over retrieval, prompt formatting, and tool calling, yet require significant engineering effort to harden for production use. Similarly, Gorilla \citep{patil2023gorilla} and Toolformer \citep{schick2023toolformer} showcase autonomous agent behaviors and tool use but leave concerns like document-level traceability, permission management, and domain adaptation to the end user.

Commercial verticals have also produced domain-specific solutions, such as ChatLaw \citep{gao2023chatlaw} for legal QA and Lawyer-LLM \citep{fan2023lawyerllm} for Chinese regulatory domains. While impressive in scope, these systems are often single-purpose and difficult to generalize beyond their training scope.

\textbf{eSapiens} differentiates itself as a full-stack platform that addresses these gaps directly.
Crucially, its architecture is engineered to unify retrieval across both unstructured sources (e.g., PDFs, Word documents, TXT files, SharePoint) and structured databases (e.g., PostgreSQL, Snowflake, RDS). This capability provides a holistic view of enterprise knowledge, a feature often requiring complex, bespoke integration in other frameworks. The platform offers hybrid vector retrieval, chunk-aware document indexing, and integrated prompt orchestration via LangChain, all encapsulated in a no-code UI layer and secured through role-based access control and tenant isolation.
Compared to open frameworks and vertical tools, eSapiens emphasizes auditable, modular, and reusable AI infrastructure that reduces integration cost, accelerates time-to-value, and meets the security and governance standards required in finance, healthcare, and legal services.
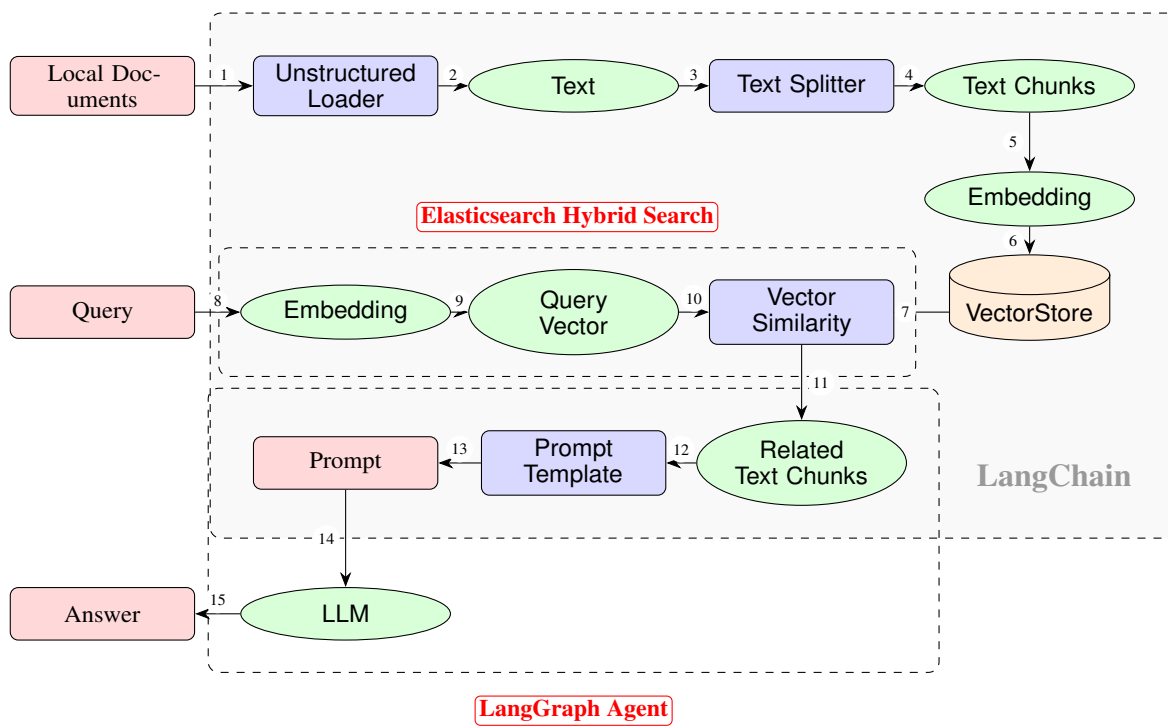
\begin{figure*}[htbp]
    \centering
    \small 

    \tikzset{
        process/.style={
            rectangle, 
            draw, 
            fill=blue!15, 
            rounded corners=3pt, 
            minimum height=2.2em,
            text width=7em,
            text centered,
            font=\sffamily\small
        },
        io/.style={
            rectangle, 
            draw, 
            fill=red!15, 
            rounded corners=3pt,
            minimum height=2.2em,
            text width=7em,
            text centered
        },
        storage/.style={
            cylinder, 
            shape border rotate=90, 
            draw, 
            fill=orange!15, 
            minimum height=2.2em,
            text width=6em,
            text centered,
            font=\sffamily\small,
            aspect=0.25
        },
        data/.style={
            ellipse, 
            draw, 
            fill=green!15, 
            minimum height=2.2em, 
            text width=5.5em,
            text centered,
            font=\sffamily\small
        },
        arrow/.style={
            draw, 
            -{Stealth[length=2mm, width=1.5mm]}
        },
        group/.style={
            draw,
            dashed,
            rounded corners,
            inner sep=8pt
        },
        grouplabel/.style={
            fill=white,
            draw=red,
            text=red,
            font=\small\bfseries,
            rectangle,
            rounded corners=2pt,
            inner sep=1.5pt
        }
    }

    \begin{tikzpicture}[node distance=1.2cm and 1.4cm]

        \node (local_docs) [io, at={(0, 9)}] {Local Documents};
        \node (loader) [process, at={(3.2, 9)}] {Unstructured Loader};
        \node (text) [data, at={(6.2, 9)}] {Text};
        \node (splitter) [process, at={(9.2, 9)}] {Text Splitter};
        \node (chunks) [data, at={(12.2, 9)}] {Text Chunks};
        
        \node (embedding1) [data, at={(12.2, 7.5)}] {Embedding};
        \node (vector_store) [storage, at={(12.2, 6)}] {VectorStore};
        
        \node (query_box) [io, at={(0, 6)}] {Query};
        \node (embedding2) [data, at={(3.2, 6)}] {Embedding};
        \node (query_vector) [data, at={(6.2, 6)}] {Query Vector};
        \node (similarity) [process, at={(9.2, 6)}] {Vector Similarity};
        
        \node (related_chunks) [data, at={(9.2, 4)}] {Related Text Chunks};
        \node (template) [process, at={(6.2, 4)}] {Prompt Template};
        
        \node (prompt) [io, at={(3.2, 4)}] {Prompt};
        \node (llm) [data, at={(3.2, 2)}] {LLM};
        \node (answer) [io, at={(0, 2)}] {Answer};

        \begin{scope}[every node/.style={font=\tiny, fill=white, circle, inner sep=1pt}]
            \path [arrow] (local_docs) -- node[above] {1} (loader);
            \path [arrow] (loader) -- node[above] {2} (text);
            \path [arrow] (text) -- node[above] {3} (splitter);
            \path [arrow] (splitter) -- node[above] {4} (chunks);
            \path [arrow] (chunks) -- node[left, xshift=-2pt] {5} (embedding1);
            \path [arrow] (embedding1) -- node[left, xshift=-2pt] {6} (vector_store);
            \path [arrow] (vector_store) -- node[left, xshift=-2pt] {7} (similarity);
            \path [arrow] (query_box) -- node[above] {8} (embedding2);
            \path [arrow] (embedding2) -- node[above] {9} (query_vector);
            \path [arrow] (query_vector) -- node[above] {10} (similarity);
            \path [arrow] (similarity) -- node[right, xshift=2pt] {11} (related_chunks);
            \path [arrow] (related_chunks) -- node[above] {12} (template);
            \path [arrow] (template) -- node[above] {13} (prompt);
            \path [arrow] (prompt) -- node[left, xshift=-2pt] {14} (llm);
            \path [arrow] (llm) -- node[above] {15} (answer);
        \end{scope}

        \begin{pgfonlayer}{background}
            
            \node[group, 
                  fit=(loader) (chunks) (vector_store) (query_vector) (template) (prompt), 
                  inner sep=16pt, 
                  fill=gray!5] (langchain_box) {};
            \node[anchor=south east, font=\large\bfseries, text=gray!80, xshift=-5mm, yshift=5mm] at (langchain_box.south east) {LangChain};
            
            \node[group, 
                  fit=(embedding2) (query_vector) (similarity), 
                  label={[grouplabel, yshift=2mm]above:Elasticsearch Hybrid Search}] (es_box) {};
            
            \node[group, 
                  fit=(related_chunks) (template) (prompt) (llm), 
                  inner sep=12pt, 
                  label={[grouplabel, yshift=-3mm]below:LangGraph Agent}] (agent_box) {};

        \end{pgfonlayer}

    \end{tikzpicture}
    
    \caption{High-level architecture of the DEREK module, illustrating the RAG pipeline from document ingestion to answer generation. The process is divided into four main stages: (1) Document Processing, (2) Query Processing, (3) Prompt Generation, and (4) Answer Generation, orchestrated within the LangChain framework and utilizing Elasticsearch for hybrid search and LangGraph for agentic verification.}
    \label{fig:derek_architecture_aligned}
\end{figure*}

\section{System Architecture}
\label{sec:architecture}
\subsection{Overview}
The DEREK module converts heterogeneous enterprise knowledge—spanning both unstructured documents (e.g., PDFs, Word files) and structured databases (e.g., SQL tables)—into fully-cited answers. As visualized for the document path in Figure \ref{fig:derek_architecture_aligned}, the pipeline has four stages. First, the \emph{Ingestion \& Chunking} stage normalises source content into retrievable units. Second, the \emph{Indexing} stage embeds each unit and stores it in a hybrid HNSW + BM25 index. Third, the \emph{Retrieval \& Reranking} stage refines the user query, runs a hybrid search, and reranks the top results. Finally, the \emph{Generation \& Verification} stage prompts the LLM and then verifies that every sentence in the draft answer is supported by a cited snippet before returning the result. The remainder of this section details each stage.

\subsection{Pipeline Stages}
\label{sec:pipeline_stages}

\paragraph{Stage 1: Ingestion \& Chunking.}
The ingestion layer is designed for data heterogeneity, handling both unstructured and structured sources through distinct but parallel processing paths. For unstructured data like DOCX, TXT, and PDF files, the module uses format-specific LangChain loaders, segmenting documents into 1,000-token chunks with a 150-token overlap. For structured data from databases like PostgreSQL or Snowflake, specialized connectors serialize table rows into natural-language sentences or structured text snippets (e.g., JSON). This unified strategy converts all knowledge into a consistent, searchable format for the subsequent indexing stage.

\paragraph{Stage 2: Indexing.}
Each text chunk is embedded using OpenAI's `text-embedding-3-large` model, with requests processed in batches for efficiency. The resulting vectors are indexed in Elasticsearch 8.x, which provides a hybrid storage solution combining a Hierarchical Navigable Small World (HNSW) graph for vector search and a parallel BM25 shard for keyword-based filtering. Both the embedding vectors and the raw text of each chunk are stored together to facilitate direct citation and verification.

\paragraph{Stage 3: Retrieval \& Reranking.}
Upon receiving a user query, the system first refines it using GPT-4o to expand synonyms and resolve ambiguity. This enhanced query is then executed against the hybrid index to retrieve 200 candidate passages. These candidates are subsequently reranked using Cohere's `rerank-english-v3` model, which rescores the passages for relevance, retaining the top-50 most relevant snippets for the generation stage.

\paragraph{Stage 4: Generation \& Verification.}
The final stage employs the CO-STAR (Context, Objective, Style, Tone, Audience, Response) prompt engineering template to guide the LLM in generating a draft answer. This draft then undergoes a second-pass verification loop managed by LangGraph. The verifier ensures that every sentence in the draft is directly supported by the retrieved snippets. If any claim fails this check, the answer is regenerated with refined instructions until all statements are grounded in the source material before being returned to the user.

\section{Experimental Results and Case Study}
\label{sec:evaluation}
We evaluate the DEREK module through a series of quantitative benchmarks and a qualitative case study in a high-stakes enterprise domain.

\subsection{Retrieval and Generation Performance}
To validate our architectural choices, we conducted two sets of experiments detailed in Appendices \ref{appendix:retrieval} and \ref{appendix:trace}.

\paragraph{Retrieval on LegalBench.} As shown in Appendix \ref{appendix:retrieval}, we benchmarked retrieval performance on four subsets of LegalBench. Our default 1,000-token chunk size improved overall Recall@50 by approximately 1 percentage point compared to a 500-token baseline (52.54\% vs. 51.82\%). The combination of hybrid search and reranking further boosts precision, demonstrating the effectiveness of our retrieval pipeline.

\paragraph{Generation Quality with TRACe.} In Appendix \ref{appendix:trace}, we used the TRACe framework to evaluate the quality of generated answers. The eSapiens pipeline, particularly when paired with models like GPT-4o and Gemini 1.5 Pro, achieved high context utilization ($\approx$0.52) and user-rated accuracy ($\geq$4.0/5.0). The built-in verifier loop was critical in minimizing unsupported claims to below 3\% in strict-grounding mode.

These results confirm that the four-stage pipeline satisfies the accuracy and traceability objectives outlined in Section~\ref{sec:motivation}.

\subsection{Case Study: Venture Capital Due Diligence}

\subsubsection{Problem Scenario}
Venture capital (VC) teams must evaluate startups quickly and rigorously. Due diligence is typically manual, repetitive, and time-consuming, with analysts spending up to \textbf{80\%} of their time on non-core tasks. A single deal may consume \textbf{100–200+} hours over several weeks, as critical documents are scattered across emails, shared drives, and multiple SaaS platforms. This creates information silos and slows down decision cycles.

\subsubsection{Solution Implementation}
We deployed the DEREK module to automate document analysis for a partner VC firm. The system ingested pitch decks, financial statements, and market reports into a single, queryable knowledge base, from which it could generate drafts of investment memos and technical reviews. This created a unified, searchable, and auditable hub for all diligence-related artifacts and AI-generated insights, streamlining the entire workflow.

\subsubsection{Results and Impact}
The AI-driven workflow significantly reduced the manual effort required for due diligence, as summarized in Table~\ref{tab:time_savings_full}. The primary outcomes were a significant acceleration in evaluation times, reducing insight generation from weeks to hours, and a decrease in the manual burden of reading and summarizing documents by over 50\%. Consequently, the firm reported closing deals 25–40\% faster and evaluating up to 10x more opportunities with the same team size. Furthermore, confidence in decisions improved by 20-30\%, as every insight was backed by a transparent source, minimizing oversight-related errors. This case study demonstrates the module's ability to deliver measurable value in a complex, real-world scenario, confirming its readiness for enterprise deployment.

\begin{table*}[htbp]
\centering
\renewcommand{\arraystretch}{1.2}
\setlength{\tabcolsep}{6pt}
\begin{tabularx}{\textwidth}{@{}
  >{\raggedright\arraybackslash}p{0.33\textwidth}
  >{\centering\arraybackslash}X
  >{\centering\arraybackslash}X
  >{\centering\arraybackslash}X@{}}
\toprule
\textbf{Area} & \textbf{Traditional Workflow} & \textbf{With AI Agent} & \textbf{Time Saved} \\
\midrule
Document review & 20--50 hrs & 5--10 mins/query & $\downarrow$\,80--90\% \\
Cohort or KPI analysis & 1.5 hrs & $<$5 mins & $\downarrow$\,95\% \\
Technical or business model review & 4--6 hrs & 15--20 mins & $\downarrow$\,90\% \\
Partner follow-up memos & 20--30 mins each & Instant & $\downarrow$\,90\%+ \\
\bottomrule
\end{tabularx}
\caption{Time savings achieved by the AI agent in the VC due diligence case study.}
\label{tab:time_savings_full}
\end{table*}

\section{Conclusion and Future Work}
The evidence presented across retrieval benchmarks, TRACe evaluations, and a real-world VC workflow confirms that the \textbf{DEREK Module} meets its design goals of accuracy, citation fidelity, and security.

To keep this advantage while scaling to larger corpora and stricter latency targets, our future work will focus on three main streams. First, we will implement sentence-aware adaptive chunking and sliding windows that expand or shrink context based on local token density and section headings, which is expected to raise Recall@50 by 1–2 pp without a latency penalty. Second, we will replace the single-stage Cohere reranker with a lightweight dual-encoder pre-filter followed by a more powerful cross-encoder for the top-100 hits, a change projected to cut rerank costs by approximately 40\% while maintaining Precision@10. Finally, we will introduce retrieval-aware prompt compression, using token-level saliency scoring to prune low-value sentences before context is sent to the LLM, enabling longer answers under the same context window and reducing spend by 15–25\%.

By executing these work packages, we aim to extend our RAG pipeline from a solid baseline to a continuously self-optimising retrieval stack capable of handling multi-terabyte corpora and sub-second latency SLAs.

\section{Limitations}
While the DEREK module demonstrates strong performance, it is subject to several inherent limitations of the RAG paradigm. First, the quality of the final answer is fundamentally capped by the quality of the retrieval stage; if relevant documents are not retrieved, the generator cannot provide a correct, cited answer. Second, the multi-step pipeline—including embedding, hybrid search, reranking, and verification—introduces computational latency and cost, particularly when using multiple external model APIs. Finally, although our verification loop significantly reduces hallucinations by enforcing citation overlap, it may not capture subtle misinterpretations or syntheses that are logically sound but not explicitly stated in the source text.

\bibliography{references}

\begin{thebibliography}{8}
\providecommand{\natexlab}[1]{#1}

\bibitem[{Chase(2022)}]{chase2022langchain}
Harrison Chase. 2022.
\newblock Langchain.
\newblock \url{https://github.com/langchain-ai/langchain}.

\bibitem[{Fan et~al.(2023)Fan, Wang, Zhang, Liu, Zhou, and Wu}]{fan2023lawyerllm}
Wei Fan, Yue Wang, Yang Zhang, Quanzhi Liu, Ao~Zhou, and Siyuan Wu. 2023.
\newblock Lawyer llm: An expert-level chinese legal large language model.
\newblock \emph{arXiv preprint arXiv:2310.10472}.

\bibitem[{Gao et~al.(2023)Gao, Han, Zhang, Zhu, He, Xu, Tang, Fu, Qian, Yu et~al.}]{gao2023chatlaw}
Yixuan Gao, Jiale Han, Yitong Zhang, Yisong Zhu, Hedan He, Yunlong Xu, Qingyan Tang, Yixuan Fu, Chao Qian, Yiming Yu, and 1 others. 2023.
\newblock Chatlaw: Open-source legal large language model with reduced hallucination.
\newblock \emph{arXiv preprint arXiv:2306.16092}.

\bibitem[{Izacard and Grave(2021)}]{izacard2021distilling}
Gautier Izacard and Edouard Grave. 2021.
\newblock Distilling knowledge from reader to retriever for question answering.
\newblock In \emph{International Conference on Learning Representations}.

\bibitem[{Lewis et~al.(2020)Lewis, Perez, Piktus, Petroni, Karpukhin, Nogueira, Paux, Genthial, Chen, Yih et~al.}]{lewis2020retrieval}
Patrick Lewis, Ethan Perez, Aleksandara Piktus, Fabio Petroni, Vladimir Karpukhin, Gustavo Nogueira, Heinrich Paux, Guillaume Genthial, Jane Chen, Wen-tau Yih, and 1 others. 2020.
\newblock Retrieval-augmented generation for knowledge-intensive nlp tasks.
\newblock In \emph{Advances in Neural Information Processing Systems}, volume~33, pages 9459--9474.

\bibitem[{Liu(2022)}]{liu2022llamaindex}
Jerry Liu. 2022.
\newblock Llamaindex.
\newblock \url{https://github.com/run-llama/llama_index}.

\bibitem[{Patil et~al.(2023)Patil, Glish, Gode, Chopite, Underscore, Underscore, and Underscore}]{patil2023gorilla}
Shishir~G Patil, Tianjun Glish, Siddesh Gode, Chaitanya Chopite, Ishan Underscore, Hamza Underscore, and Raj Underscore. 2023.
\newblock Gorilla: Large language model connected with massive apis.
\newblock \emph{arXiv preprint arXiv:2305.15334}.

\bibitem[{Schick et~al.(2023)Schick, Dwivedi-Yu, Dess{\`\i}, Raileanu, Lomeli, Zettlemoyer, Cancedda, and Scialom}]{schick2023toolformer}
Timo Schick, Jane Dwivedi-Yu, Roberto Dess{\`\i}, Roberta Raileanu, Maria Lomeli, Luke Zettlemoyer, Nicola Cancedda, and Thomas Scialom. 2023.
\newblock Toolformer: Language models can teach themselves to use tools.
\newblock \emph{arXiv preprint arXiv:2302.04761}.

\end{thebibliography}

\appendix
\section*{Appendix}
\section{Retrieval Performance on Long-form Legal QA}
\label{appendix:retrieval}

\subsection{Objectives and Scope}
This benchmark evaluates the DEREK pipeline on long-form legal QA tasks, with a focus on four datasets. It aims to answer three core questions regarding retrieval coverage, the precision-recall tradeoff, and scalability. Specifically, we investigate how often the top-k retrieved passages fully cover the answer span, determine which chunk size (500 vs. 1000 tokens) better balances completeness and retrieval accuracy, and assess which configuration is most suitable for real-world workloads.

\subsection{Datasets}
The LegalBench subsets used in the benchmark are detailed in Table \ref{tab:datasets}.

\begin{table}[H]
\centering
\small
\renewcommand{\arraystretch}{1.2} 
\setlength{\tabcolsep}{4pt} 
\begin{tabular}{@{}l*{4}{S[table-format=4.0]}@{}}
\toprule
\textbf{Metric} & \textbf{PrivacyQA} & \textbf{CUAD} & \textbf{MAUD} & \textbf{ContractNLI} \\
\midrule
Q--A Pairs & 712 & 1012 & 1215 & 3950 \\
\bottomrule
\end{tabular}
\caption{LegalBench subsets used in the benchmark.}
\label{tab:datasets}
\end{table}

\subsection{Quantitative Results: Chunk Comparison}
The Table 3 and Table 4 compare Recall and Precision scores for \textit{eSapiens} under two representative chunk windows—\textbf{500 tokens}, the default size used by many open-source RAG toolkits, and \textbf{1000 tokens}, the larger window shipped as our module’s out-of-the-box setting to capture more intra-paragraph context—across all four datasets.

\begin{table*}[htbp]
\centering
\small
\renewcommand{\arraystretch}{1.1}
\setlength{\tabcolsep}{5pt}
\begin{tabular}{lcccccc|cccccc}
\toprule
\textbf{Dataset} & \multicolumn{6}{c|}{\textbf{Recall@k (\%)}} & \multicolumn{6}{c}{\textbf{Precision@k (\%)}} \\
\cmidrule(lr){2-7} \cmidrule(lr){8-13}
 & k=1 & k=2 & k=4 & k=8 & k=16 & k=50 & k=1 & k=2 & k=4 & k=8 & k=16 & k=50 \\
\midrule
PrivacyQA   & 18.15 & 25.87 & 49.28 & 64.07 & 85.63 & 96.47 & 18.50 & 14.02 & 13.18 & 9.26 & 4.74 & 5.28 \\
ContractNLI & 4.91  & 9.33  & 16.09 & 25.83 & 35.04 & 46.90 & 5.08  & 5.59  & 5.04  & 3.67 & 2.52 & 1.75 \\
MAUD        & 0.52  & 2.48  & 4.39  & 7.24  & 14.03 & 22.60 & 1.94  & 2.63  & 2.05  & 1.77 & 1.79 & 1.12 \\
CUAD        & 3.17  & 7.33  & 18.26 & 28.67 & 42.50 & 55.66 & 3.53  & 4.18  & 6.18  & 5.06 & 3.93 & 2.74 \\
\midrule
\textbf{ALL} & \textbf{7.26} & \textbf{11.52} & \textbf{20.40} & \textbf{27.94} & \textbf{41.37} & \textbf{51.82} & \textbf{7.49} & \textbf{6.82} & \textbf{6.65} & \textbf{5.02} & \textbf{3.78} & \textbf{2.68} \\
\bottomrule
\end{tabular}
\caption{Performance of eSapiens (Chunk = 500): Recall and Precision at Different $k$.}
\label{tab:chunk500}
\end{table*}

\begin{table*}[htbp]
\centering
\small
\renewcommand{\arraystretch}{1.1}
\setlength{\tabcolsep}{5pt}
\begin{tabular}{lcccccc|cccccc}
\toprule
\textbf{Dataset} & \multicolumn{6}{c|}{\textbf{Recall@k (\%)}} & \multicolumn{6}{c}{\textbf{Precision@k (\%)}} \\
\cmidrule(lr){2-7} \cmidrule(lr){8-13}
 & k=1 & k=2 & k=4 & k=8 & k=16 & k=50 & k=1 & k=2 & k=4 & k=8 & k=16 & k=50 \\
\midrule
PrivacyQA   & 10.10 & 20.24 & 28.84 & 54.95 & 71.44 & 94.47 & 8.97  & 10.31 & 7.81  & 7.34  & 5.16 & 2.64 \\
ContractNLI & 4.81  & 8.72  & 12.62 & 17.85 & 25.54 & 39.78 & 2.28  & 2.47  & 1.84  & 1.33  & 0.89 & 0.42 \\
MAUD        & 0.52  & 2.48  & 3.05  & 4.57  & 7.31  & 13.60 & 1.33  & 1.07  & 0.84  & 0.64  & 0.53 & 0.32 \\
CUAD        & 3.62  & 10.47 & 20.63 & 32.46 & 45.24 & 62.30 & 2.12  & 3.08  & 3.17  & 2.70  & 2.01 & 0.96 \\
\midrule
\textbf{ALL} & \textbf{4.93} & \textbf{10.34} & \textbf{16.29} & \textbf{27.46} & \textbf{37.38} & \textbf{52.54} & \textbf{3.68} & \textbf{4.23} & \textbf{3.42} & \textbf{3.00} & \textbf{2.15} & \textbf{1.09} \\
\bottomrule
\end{tabular}
\caption{Performance of eSapiens (Chunk = 1000): Recall and Precision at Different $k$.}
\label{tab:chunk1000}
\end{table*}

\begin{table*}[htbp]
\centering
\small
\renewcommand{\arraystretch}{1.1}
\setlength{\tabcolsep}{5pt}
\begin{tabular}{lccccc}
\toprule
\textbf{Model Pipeline} & \textbf{Completeness} & \textbf{Utilization} & \textbf{Context Rel.} & \textbf{Hallucination} & \textbf{Accuracy} \\
\midrule
\multicolumn{6}{l}{\textit{eSapiens DEREK Pipeline}} \\
\quad + GPT-4o & 0.4307 & 0.5224 & 0.3648 & 0.1823 & 3.85 \\
\quad + GPT-4o-mini & 0.3433 & 0.4658 & 0.3785 & 0.2729 & 3.70 \\
\quad + Claude 3.7 & 0.3840 & 0.4440 & 0.3648 & 0.1403 & 4.05 \\
\quad + Gemini 1.5 Pro & 0.4982 & 0.5179 & 0.3728 & 0.1712 & 4.00 \\
\quad + DeepSeek-R1 & 0.4571 & 0.5167 & 0.2581 & 0.1486 & 4.15 \\
\midrule
\multicolumn{6}{l}{\textit{Baseline FAISS Pipeline (top-2, short prompt)}} \\
\quad + GPT-4o & 0.4450 & 0.4800 & 0.3294 & 0.0875 & 3.55 \\
\quad + GPT-4o-mini & 0.4105 & 0.4467 & 0.3090 & 0.1524 & 3.15 \\
\quad + Claude 3.7 & 0.4985 & 0.5102 & 0.3270 & 0.0860 & 3.75 \\
\quad + Gemini 1.5 Pro & 0.5346 & 0.5728 & 0.3296 & 0.1371 & 3.75 \\
\quad + DeepSeek-R1 & 0.5381 & 0.5102 & 0.3430 & 0.1139 & 4.15 \\
\bottomrule
\end{tabular}
\caption{Evaluation results of DEREK models on 100 random questions from RAGtruth.}
\label{tab:trace_results}
\end{table*}

\subsection{Analysis of Chunking Strategy}
\noindent\textbf{PrivacyQA:} Due to fragmented original text, chunk = 500 yields higher recall at low $k$, but performance converges with chunk = 1000 as $k$ increases.

\noindent\textbf{Other datasets:} For \textit{CUAD}, \textit{MAUD}, and \textit{ContractNLI}, chunk = 1000 consistently outperforms 500 in recall, especially at top-50. Larger chunks preserve more semantic information.

\noindent\textbf{Precision:} While more affected by chunk size, precision is less critical than recall in business applications. Strong models like GPT-4o can filter irrelevant chunks.

\noindent\textbf{Real-world usage:} Chunk = 1000 is better aligned with production needs, offering broader coverage and lower information loss.

\subsection{Summary of Findings}
Through comparative experiments on four datasets — \textit{PrivacyQA}, \textit{CUAD}, \textit{MAUD}, and \textit{ContractNLI} — clear trends are observed in how chunk window size affects retrieval performance.

\noindent\textbf{PrivacyQA:}
Due to the fragmented nature of its original text, \textit{PrivacyQA} benefits from smaller window sizes. With chunk = 500, vector search achieves higher location precision, leading to superior Recall@k at low $k$ values. However, as $k$ increases, the recall performance of both 500 and 1,000 converge.

\noindent\textbf{Other datasets:}
For \textit{CUAD}, \textit{MAUD}, and \textit{ContractNLI}, chunk = 1,000 consistently outperformed 500 in recall, especially at high $k$ values (e.g., $k=50$). This shows that larger chunks better preserve semantic context and retrieve more useful information.

\noindent\textbf{Precision considerations:}
Although precision varied more significantly across chunk sizes, it is found that precision is less critical than recall in practical workflows. Given the reasoning capabilities of large models such as GPT‑4o, low-precision context can still be filtered intelligently.

\noindent\textbf{Real-world usage:}
In enterprise applications where completeness and recall are paramount, chunk = 1,000 proves more advantageous. While chunk = 500 may slightly improve precision, chunk = 1,000 delivers a better recall–quality tradeoff.

\section{TRACe-Based Output Quality Evaluation}
\label{appendix:trace}
\subsection{Objectives and Setup}
This experiment evaluates the generation quality of eSapiens and a baseline FAISS-based RAG pipeline using the TRACe framework. Five popular LLMs (GPT-4o, GPT-4o-mini, Claude 3.7, Gemini 1.5 Pro, DeepSeek R1) are tested under identical question sets. Key evaluation dimensions include completeness (whether key answer points were reflected), utilization (proportion of retrieved context used), context relevance (alignment with user intent), hallucination (percent of unsupported tokens), and accuracy (human-graded alignment with ground truth).

\subsection{Results Overview}
Table~\ref{tab:trace_results} summarizes the TRACe evaluation metrics for each model across the eSapiens and FAISS pipelines.

\subsection{Findings and Analysis}
First, regarding hallucination, the FAISS baseline shows fewer instances than eSapiens. This is explained by its strict generation constraint: FAISS forces the LLM to answer only with retrieved text, returning “I don’t know” if evidence is missing. The default eSapiens workflow allows limited abstraction, which can occasionally introduce unsupported content. Within eSapiens, this can be mitigated by enabling the built-in *strict-grounding* preset, which adds a rule to answer only from context, tightens the citation verifier, and allows a "refuse-to-answer" path. This preset lowers hallucination risk at the cost of a slightly higher refusal rate, a trade-off suitable for compliance-critical scenarios.

Second, the FAISS baseline yields higher completeness scores. By constraining generation to the top-k retrieved passages, FAISS ensures dense coverage of the provided context. In contrast, eSapiens sometimes applies structural formatting or abstraction, which can lead to the partial omission of some points from the source snippets.

Third, eSapiens outperforms FAISS in context relevance. Even when retrieving the same number of documents, the eSapiens pipeline uses hybrid retrieval and reranking strategies that produce more targeted, intent-aligned context for the generator.

Fourth, regarding context utilization, models like Gemini-1.5-pro and GPT-4o perform better on the eSapiens pipeline. These models are particularly strong at reasoning over the well-structured paragraphs provided by eSapiens, with Gemini excelling at structured summarization and GPT-4o at direct textual citation.

Finally, in terms of human-judged accuracy and fluency, eSapiens achieves higher scores. Its integrated post-processing, which includes formatting and rephrasing, improves the readability and coherence of the final answer while preserving the core meaning.

\subsection{Summary of Findings}
In summary, eSapiens shows clear gains in relevance, fluency, and model alignment—particularly when paired with Gemini or GPT-4o. The FAISS baseline yields stricter factual alignment with fewer hallucinations but lacks the naturalness and flexibility found in eSapiens generations. Each pipeline offers tradeoffs that may suit different use cases (e.g., regulatory Q\&A vs. creative summarization).

\end{document}